\documentclass{bmvc2k}

\usepackage{amssymb}
\usepackage{multirow}
\title{ITC-RWKV: Interactive Tissue–Cell Modeling with Recurrent Key-Value Aggregation for Histopathological Subtyping}

\addauthor{Yating Huang$^*$}{yating.huang@manchester.ac.uk}{1}
\addauthor{Qijun Yang$^*$}{qijun.Yang@manchester.ac.uk}{1}
\addauthor{Lintao Xiang}{ltxiang.work@gmail.com}{1}
\addauthor{Hujun Yin$^\dagger$}{hujun.yin@manchester.ac.uk}{1}

\addinstitution{
 Department of Electrical and Electronic Engineering\\
 The University of Manchester\\
 Manchester, UK
}

\runninghead{Huang et al.}{ITC-RWKV for Tissue–Cell Modeling and Subtyping}

\begin{document}
\makeatletter
\renewcommand\thefootnote{}
\makeatother
\maketitle
\footnotetext{%
$^*$Equal contribution.\quad $^\dagger$Corresponding author.}

\begin{abstract}
Accurate interpretation of histopathological images demands integration of information across spatial and semantic scales, from nuclear morphology and cellular textures to global tissue organization and disease-specific patterns. Although recent foundation models in pathology have shown strong capabilities in capturing global tissue context, their omission of cell-level feature modeling remains a key limitation for fine-grained tasks such as cancer subtype classification. To address this, we propose a dual-stream architecture that models the interplay between macroscale tissue features and aggregated cellular representations. 
To efficiently aggregate information from large cell sets, we propose a receptance-weighted key-value aggregation model, a recurrent transformer that captures inter-cell dependencies with linear complexity. Furthermore, we introduce a bidirectional tissue-cell interaction module to enable mutual attention between localized cellular cues and their surrounding tissue environment. Experiments on four histopathological subtype classification benchmarks show that the proposed method outperforms existing models, demonstrating the critical role of cell-level aggregation and tissue-cell interaction in fine-grained computational pathology.
\end{abstract}

%-------------------------------------------------------------------------
\section{Introduction}
\label{sec:intro}
Histopathological examination of biopsy specimens remains a cornerstone of cancer diagnosis, providing detailed insights into tissue architecture and cellular abnormalities \cite{tosta2019computational}. Early tumor detection and accurate subtyping are pivotal for optimizing treatment decisions and improving patient survival \cite{Allemani2015_CONCORD2}. However, traditional workflows that rely on expert microscopic review of stained tissue sections are time-consuming and prone to inter-observer variability and diagnostic error. The growing adoption of digital pathology, along with advances in computer vision, has enabled the development of computational frameworks for automated histological analysis \cite{Campanella2019_CompPathology, Chen2024_UNI, Kather2019_MSI, Lu2021_CLAM, Lu2024_CONCH}. Deep learning based models have shown strong potentials in accelerating the diagnostic workflows and offering more consistent and scalable assessments across diverse clinical contexts.

However, unlocking the full diagnostic potential of digital pathology presents several key challenges. First, pathological diagnosis by experts often hinges on fine-grained nuclear morphology, including variations in size, shape, and chromatin distribution. While accurate nuclear segmentation and individual cell feature encoding can significantly boost diagnostic performance \cite{graham2019hover, horst2024cellvit}, most existing multiple instance learning (MIL) or Transformer-based methods process entire image patches as the minimal unit, inevitably diluting crucial cellular signals through global averaging or max pooling operations \cite{shao2021transmil}. 
In contrast, cell graph networks explicitly model inter-nuclear relationships but introduce substantial computational overhead. In dense regions, they may generate thousands of nodes, raising scalability issues \cite{pati2020hact, chen2021patchgcn}. 
This leads to a second major challenge: how to efficiently aggregate features from potentially hundreds or even thousands of cells within a patch region while preserving their individual characteristics. Traditional self-attention scales quadratically with the number of tokens ($\mathcal{O}(n^2)$) and hence are prohibitively expensive for cell-rich fields of view. Although approximate variants like Set Transformer \cite{lee2019set} and LINformer \cite{lee2019set} offer dimensionality reduction, they still encounter memory bottlenecks on long sequences due to global attention requirements. 
Finally, precise diagnosis of complex cases such as ductal carcinoma in situ (DCIS) versus invasive carcinoma (IC) 
depends on identifying subtle cell–tissue interface events, like basement membrane breach. Similarly, assessing tumor-infiltrating lymphocytes for prognosis or immunotherapy requires modeling both cellular identity and spatial distribution within the microenvironment. 
These tasks demand models that can embed nuclear features in tissue context and support bidirectional communication between cell-level and tissue-level representations to enhance diagnostic accuracy and interpretability.
% while enabling bidirectional communication between cell-level and tissue-level representations to enhance diagnostic accuracy and interpretability.
%These tasks demand modeling strategies that contextualize cell-level information within surrounding tissue and support meaningful interaction across spatial scales, ultimately improving the accuracy and interpretability of computational pathology. 
%Therefore, developing modules capable of precisely aligning cell-level information with their peritumoral environment and facilitating bidirectional information flow between cellular and tissue representations is essential for improving the accuracy and interpretability of digital pathology diagnostics.

To address these challenges, we propose a novel Interactive Tissue–Cell Network with RWKV aggregation (ITC-RWKV), a dual-stream framework that synergistically integrates cellular and tissue-level information for fine-grain histopathology classification. 
The method features a dedicated cell pathway that performs instance segmentation of nuclei, extracts individual nuclear embeddings, and aggregates them using a linear-complexity mechanism based on the Receptance Weighted Key-Value architecture, which we term Aggr-RWKV. This design overcomes the limitations of traditional pooling and quadratic attention, enabling efficient modeling of large cell populations.
In parallel, a tissue pathway leverages a powerful foundation model pre-trained on diverse pathology data to encode global tissue architectural patterns. Critically, we introduce a novel tissue–cell interaction module that supports bidirectional information flow between local and global representations. It employs contextual ROI pooling to align tissue features with individual cell locations and utilizes dual cross-attention to mutually refine cellular and tissue representations, capturing crucial interplay between cells and their microenvironment.

The main contributions are threefold: 
\noindent(i) a dual-stream architecture is proposed to jointly encode cellular and tissue-level cues, faithfully mirroring the workflow of pathologists; 
\noindent(ii) we introduce \emph{Aggr-RWKV}, a linear-complexity nuclear aggregation mechanism that scales to hundreds of cells while retaining rich morphology; and 
\noindent(iii) a bidirectional tissue–cell interaction module is designed to capture micro-environmental context, boosting accuracy on challenging subtypes and yielding interpretable feature attributions.

\section{Related Work}

\noindent\textbf{Pathological Image Classification:} Deep learning methods have become widely adopted in pathological image classification, frequently utilizing a MIL framework to handle large images by processing and aggregating features from numerous components. Early approaches relied on simple aggregation techniques, such as max or mean pooling. Later, attention mechanisms, like those in AB-MIL \cite{ilse2018attentionmil,shao2021transmil,yao2020wholeattentionmil}, were introduced to better signify components based on relevance. Graph-based methods, such as Patch-GCN and graph transformer networks \cite{zhou2019cgcgnn, pati2022hierarchicalgnn,aygunecs2020graphpatchgnn}, further advanced this approach by capturing structural relationships between image components. To effectively utilize vast unlabeled pathological data and reduce annotation burden, self-supervised learning (SSL) trains models via pretext tasks, yielding powerful, generalizable representations that serve as a strong foundation for downstream MIL or graph-based analysis \cite{ciga2022self,li2021dualself,koohbanani2021self}. This capability for large-scale, efficient pre-training using unlabeled data is precisely what enables the development of the latest generation of foundation models in pathology (e.g. CTransPath ~\cite{wang2022transformer}). Models such as GPFM \cite{ma2024gpfm}, Virchow \cite{vorontsov2024virchow}, and UNI \cite{Chen2024_UNI} benefit from being pre-trained on massive datasets, thereby offering enhanced representation capabilities for downstream tasks. Despite these advancements, challenges remain in preserving fine cellular details, efficiently aggregating numerous features, and integrating cellular structural insights with broader tissue context for more accurate and interpretable diagnoses.

\noindent\textbf{Receptance Weighted Key Value (RWKV):} 
RWKV \cite{peng2023rwkv} was initially developed for natural language processing (NLP) tasks. It offers an efficient architecture that blends the efficient parallel training capabilities akin to Transformers \cite{vaswani2017attention} with the linear-time inference characteristic of RNNs. 
It addresses the quadratic complexity of standard self-attention through a WKV mechanism for processing long-range dependencies and employs a token shift for capturing local context.
Initially successful in NLP, this architecture was subsequently extended to computer vision with Vision-RWKV \cite{duan2024visionrwkv}, demonstrating efficiency advantages in handling high-resolution data. Subsequent work has further explored RWKV variants for diverse visual tasks, including Diffusion-RWKV \cite{fei2024diffusionrwkv} for image generation, RWKV-SAM \cite{yuan2024samrwkv} for segmentation, Point-RWKV \cite{he2025pointrwkv} for 3D point clouds, and RWKV-CLIP \cite{gu2024rwkvclip} for vision-language representation learning. 
Given RWKV's strengths in processing long sequences and maintaining efficiency, we explore its adaptation to medical imagery, specifically for modeling fine-grained cellular structures in histopathology. Its sequential nature and linear scalability make it well-suited for aggregating large numbers of instances while preserving spatial and morphological context.

%%%%%%%%%%%%%%%%%%%%%%%%%%%%%%%%%%%%%%%%%%%%%%%%%%%%%%%%%%%%%%%%%%%%%%%%%%%%
\section{Methodology}
\begin{figure*}
\centering
\includegraphics[width=\linewidth]{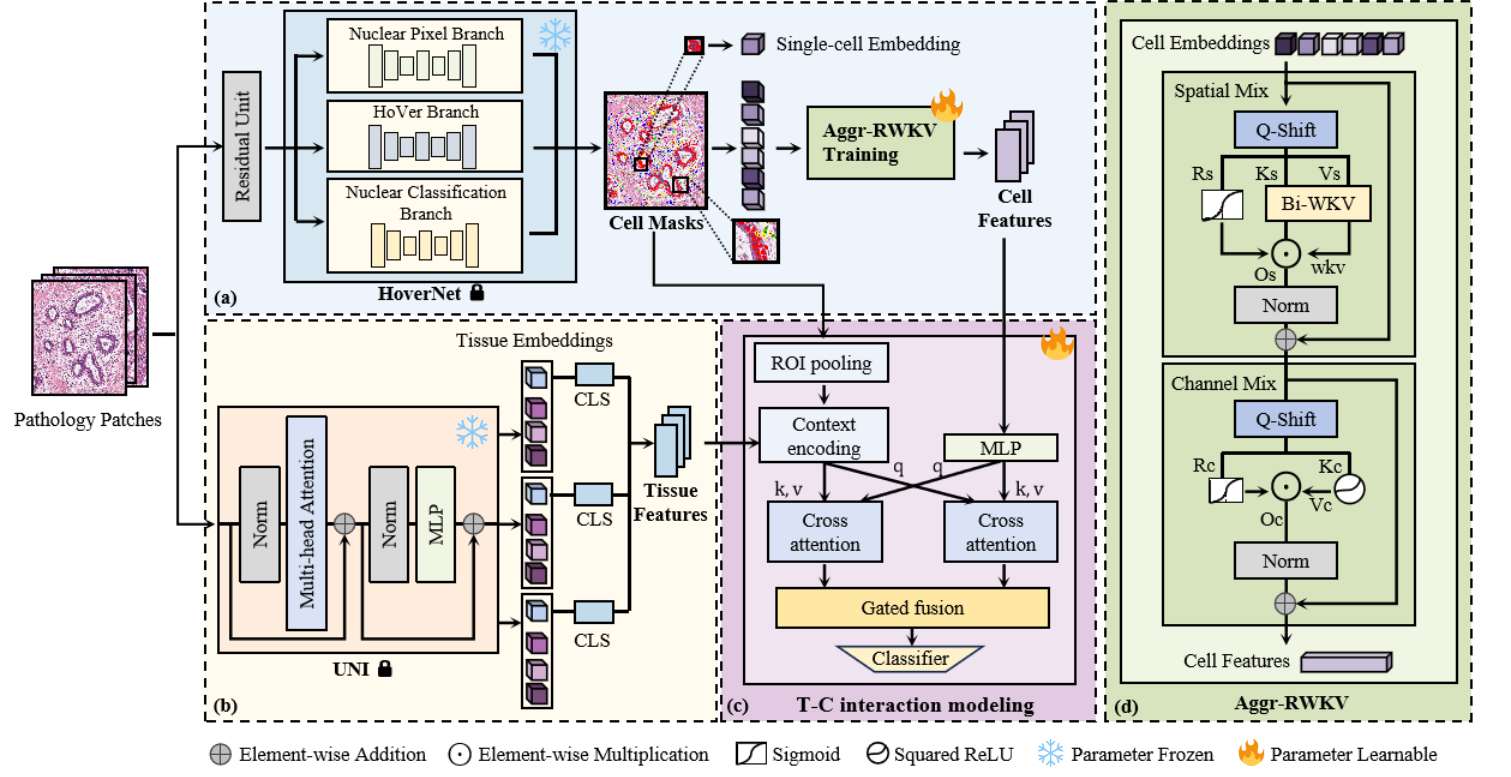}
\caption{Overview of the proposed ITC-RWKV model.}
\label{fig:fig1}
\end{figure*}
\subsection{Overall Pipeline}
Digital pathology diagnosis requires analysis across two complementary scales: individual cellular morphology and global tissue architecture. Our framework addresses this challenge through a dual-stream design that mimics the diagnostic process of pathologists. As illustrated in Figure \ref{fig:fig1}, it consists of three key components: (i) a cell pathway (Figure \ref{fig:fig1}a) that processes individual nuclei and aggregates their features using the proposed Aggr-RWKV module (Figure \ref{fig:fig1}d), (ii) a tissue pathway (Figure \ref{fig:fig1}b) leveraging the pathological foundation model for contextual understanding, and (iii) a tissue-cell interaction module (Figure \ref{fig:fig1}c) enabling cross-scale information exchange.

To start with, we employ the UNI \cite{Chen2024_UNI} foundation model to extract tissue-level features that capture architectural context. In parallel, the cell pathway processes fine-grained nuclear morphology features and aggregates them via the proposed Aggr-RWKV, a scalable recurrent transformer. These two morphology are then unified through our novel interaction mechanism that models tissue-cell interactions, creating a comprehensive representation that leverages both cellular and tissue scales. Formally, given a tissue-level pathology image $I$, our goal is to predict labels by combining these complementary views:
\begin{equation}
p(y|I) = \operatorname{softmax}(f_{\text{cls}}(\phi(\mathbf{c}, \mathbf{t}_{\text{[CLS]}})))
\end{equation}
where $\mathbf{c}$ represents aggregated cellular features, $\mathbf{t}_{\text{[CLS]}}$ encodes tissue-level information, and $\phi(\cdot,\cdot)$ is our tissue-cell interaction function that fuses information from both branches. The following sections provide detailed descriptions of the cell aggregation module (Aggr-RWKV) and the tissue–cell interaction mechanism.

\subsection{Cell Pathway with Aggr-RWKV}
The cell pathway (Figure \ref{fig:fig1}a) extracts and aggregates nuclear features to capture morphological characteristics critical for diagnosis. We first utilize HoverNet\cite{graham2019hover} for nuclear instance segmentation, which processes the input image through three specialized branches: (i) a Nuclear Pixel branch that separates nuclear pixels from background, (ii) a HoVer branch that computes horizontal and vertical distances to nucleus centers, facilitating separation of touching nuclei, and (iii) a Nuclear Classification branch that determines nucleus types. These three branches work together to produce high-quality cell masks $\{\mathcal{R}_k\}$.
For each segmented nucleus, we extract embeddings using a lightweight CNN: $\mathbf{h}_k = f_{\text{cell}}(I[\mathcal{R}_k])$. This results in a set of unordered cell descriptors $\mathcal{H} = \{\mathbf{h}_k\}$ that encode nuclear properties like size, shape, and chromatin distribution.

A key challenge in this process is efficiently aggregating these features from potentially hundreds of nuclei while preserving their morphological characteristics. 
Existing approaches either lose critical cell-specific information through simple pooling or suffer from quadratic computational complexity with self-attention. 
To address this, we introduce Aggr-RWKV (Figure \ref{fig:fig1}d), which adapts the receptance weighted key-value architecture to create an efficient aggregation mechanism.

Given the cell embedding matrix $\mathbf{H} = [\mathbf{h}_1,...,\mathbf{h}_n]^{\top}$, Aggr-RWKV consists of two core components: spatial mixing via bidirectional WKV attention and channel mixing through gated modulation.

\noindent\textbf{Spatial Mixing:}\hspace{0.8em}Inspired by \cite{duan2024visionrwkv}, this module computes bidirectional attention using the Bi-WKV mechanism. The input features $\mathbf{H}$ are first processed by Q-Shift (a quad-directional token shift that interpolates each token with its spatial neighbors to enlarge the receptive field) to obtain shifted features $\mathbf{H}_{\text{shifted}}$. As shown in Figure \ref{fig:fig1}c, three independent linear projections are then applied to generate the receptance vector $\mathbf{R}_s = \mathbf{H}_{\text{shifted}}\mathbf{W}_r$, key vector $\mathbf{K}_s = \mathbf{H}_{\text{shifted}}\mathbf{W}_k$, and value vector $\mathbf{V}_s = \mathbf{H}_{\text{shifted}}\mathbf{W}_v$. With linear complexity, the Bi-WKV mechanism recurrently aggregates $\mathbf{K}_s$ and $\mathbf{V}_s$ in both forward and backward directions, functioning like a recurrent model to capture bidirectional context and produce the output $\mathbf{wkv}$:
%With linear complexity, Bi-WKV recurrently aggregates $\mathbf{K}_s$ and $\mathbf{V}_s$ in both forward and backward directions to capture bidirectional context, producing the output $\mathbf{wkv}$:
\begin{equation}
\mathbf{wkv} = \text{Bi-WKV}(\mathbf{K}_s, \mathbf{V}_s)
\end{equation}
Finally, the module's output $\mathbf{O}_s$ is obtained by gating (element-wise multiplication) the $\mathbf{wkv}$ output with the sigmoid-activated receptance vector $\mathbf{R}_s$. This result is then passed through layer normalization and a residual connection to produce the block's final output, $\mathbf{H}_{\text{s}}$.
\begin{equation}
\mathbf{O}_s = \sigma(\mathbf{R}_s) \odot \mathbf{wkv},\quad \mathbf{H}_{\text{s}} = \mathbf{H} + \operatorname{LayerNorm}(\mathbf{O}_s)
\end{equation}

\noindent\textbf{Channel Mixing:}\hspace{0.8em}This module models intra-feature interactions from the previous output $\mathbf{H}_{\text{s}}$. The input is first processed by Q-Shift to obtain $\mathbf{H}_{\text{s\_shifted}}$, which is linearly projected into a receptance vector $\mathbf{R}_c = \mathbf{H}_{\text{s\_shifted}}\mathbf{W}_r$ and Key vector $\mathbf{K}_c = \mathbf{H}_{\text{s\_shifted}}\mathbf{W}_k$. The output $\mathbf{O}_c$ is then computed by applying a SquaredReLU activation to $\mathbf{K}_c$ and gating it with $\sigma(\mathbf{R}_c)$ through element-wise multiplication:

\begin{equation}
\mathbf{O}_c = \sigma(\mathbf{R}_c) \odot \text{SquaredReLU}(\mathbf{K}_c)
\end{equation}
This combination allows the model to capture complex relationships among the internal features of each cell representation.

The final output of the entire Aggr-RWKV block is obtained from the output of the channel mixing stage after layer normalization and a residual connection:
\begin{equation}
\mathbf{H}_{cell} = \mathbf{H}_{\text{spatial}} + \operatorname{LayerNorm}(\mathbf{O}_c)
\end{equation}
This updated feature matrix contains enhanced cell representations that incorporate inter-cell relationship information and feature-level interactions. By stacking multiple Aggr-RWKV blocks, cell representations are progressively refined and ultimately aggregated into a global cell feature vector for downstream tasks.

\subsection{Tissue-Cell Interaction Modeling}

Many diagnostic cues in pathology emerge from the spatial relationships between cells and their surrounding tissue microenvironment. To model these interactions, we design a bidirectional fusion module (Figure \ref{fig:fig1}c) that enables reciprocal refinement between cellular and tissue-level representations.

We begin by aligning spatially corresponding features from the two branches. For each nucleus with mask $\mathcal{R}_k$, we extract its contextual tissue representation $\mathbf{r}_k$ by applying ROI pooling over the token features from the tissue branch (Figure \ref{fig:fig1}b). This pooling aggregates token embeddings from the UNI model that overlap with the cell’s position and its local microenvironment. As a result, we obtain paired sequences: $\{\mathbf{h}_k\}$ from the cell pathway, and corresponding tissue contexts $\{\mathbf{r}_j\}$. To enable information exchange between the two modalities, we perform dual cross-attention:

\begin{equation}
\tilde{\mathbf{h}}_k = \operatorname{Attn}\bigl(\mathbf{h}_k,\{\mathbf{r}_j\},\{\mathbf{r}_j\}\bigr), \quad
\tilde{\mathbf{r}}_k = \operatorname{Attn}\bigl(\mathbf{r}_k,\{\mathbf{h}_j\},\{\mathbf{h}_j\}\bigr)
\end{equation}
where $\operatorname{Attn}(q, k, v)$ represents the standard attention mechanism. This mechanism allows cell representations to be enhanced by tissue context, while simultaneously enriching tissue features with cellular details. After cross-attention, we aggregate the context-enriched cell features $\{\tilde{\mathbf{h}}_k\}$ into a global cellular representation $\mathbf{c}$. Similarly, the cell-aware tissue features $\{\tilde{\mathbf{r}}_k\}$ are combined with the [CLS] token embedding $\mathbf{t}_{\text{[CLS]}}$ to form a comprehensive tissue representation. The attended outputs are then combined through our tissue-cell interaction function $\phi$, which implements a gated fusion mechanism:
\begin{equation}
\mathbf{z} = \sigma\!\bigl(\mathbf{W}_g[\mathbf{c};\mathbf{t}_{\text{[CLS]}}]\bigr)\odot \mathbf{c}
+ \bigl(1-\sigma(\mathbf{W}_g[\mathbf{c};\mathbf{t}_{\text{[CLS]}}])\bigr)\odot\mathbf{t}_{\text{[CLS]}}
\end{equation}
where $\mathbf{W}_g$ is a learnable weight matrix, $[\cdot;\cdot]$ denotes concatenation, and $\sigma$ is the sigmoid activation function. This learned gating mechanism adaptively balances cellular and tissue information based on diagnostic relevance. The final fused representation $\mathbf{z}$ serves as input to our classification head $f_{\text{cls}}$, which consists of a multi-layer perceptron with one hidden layer and dropout for regularization.

By explicitly modeling cell-tissue interactions and integrating information across scales, our approach captures complex spatial relationships considered in human diagnosis, enabling more nuanced feature representations that reflect the hierarchical organization of pathological tissues. This architecture's adaptive fusion achieves a balance between fine-grained cellular details and broader tissue patterns necessary for accurate pathological diagnosis.
%\vspace{-1pt}

\section{Experiments}
\subsection{Datasets}
We evaluated our method on four public histopathology benchmarks covering two cancer types and multiple domain shifts. 

For breast cancer, \textbf{BRACS} \cite{brancati2022bracs} contains 4,391 H\&E-stained regions of interest (RoIs) from 325 whole slide images (0.25 $\mu m$/pixel), annotated into seven diagnostic categories: Normal, Benign, Usual Ductal Hyperplasia (UDH), Atypical Ductal Hyperplasia (ADH), Flat Epithelial Atypia (FEA), Ductal Carcinoma In Situ (DCIS), and Invasive Carcinoma. We follow the official train/validation/test splits. \textbf{BACH} \cite{aresta2019bach} includes 500 high-resolution images (1536×2048 pixels, 0.42 $\mu m$/pixel) labeled into four categories: Normal, Benign, In Situ Carcinoma, and Invasive Carcinoma, providing a complementary evaluation scenario with fewer but balanced classes.

For prostate cancer, \textbf{UHU} \cite{arvaniti2018automated} comprises 22,022 image patches (750×750 pixels, 40× magnification) across benign (2,076 train / 127 test) and three Gleason grades—grade 3 (6,303 / 1,602), grade 4 (4,541 / 2,121), and grade 5 (2,383 / 387)—with the test set representing in-domain evaluation. \textbf{UBC} \cite{karimi2019deep}, from the Gleason2019 challenge, contains 7,260 patches (690×690 pixels, 40× magnification) with the same four categories. Differences in scanners, staining, and patient cohorts introduce substantial domain shifts, making UBC a challenging cross-domain test set.

This combination of datasets spans variations in cancer type, diagnostic granularity, resolution, and acquisition protocols, enabling rigorous evaluation of both in-domain accuracy and cross-domain generalization.

\subsection{Implementation Details}
Our framework was implemented in PyTorch and trained on NVIDIA A100 GPUs (40GB). The tissue pathway used the UNI vision transformer initialized with weights pretrained on a diverse collection of pathology images. For the cell pathway, we used HoverNet for cell instance segmentation, which was pretrained on PanNuke dataset \cite{gamper2020pannuke} and kept frozen during training. Each detected nucleus was processed through a lightweight CNN encoder to extract 256-dimensional cell features. We employed the Adam \cite{kingma2014adam} optimizer with a learning rate of 1e-4 for the cell encoder, with cosine learning rate decay. All models were trained with a batch size of 16 for 100 epochs, applying early stopping based on validation performance.

\subsection{Performance Comparisons}
\noindent\textbf{Main results:} We evaluated our model against state-of-the-art approaches on both BRACS and BACH datasets. Table \ref{tab:main} reports F1 scores on BRACS across diagnostic category and the weighted average.
The proposed approach achieved the highest overall F1 score of 66.5\%, significantly outperforming the previous best methods. The improvements are particularly notable for the challenging categories: normal tissues (76.3\%, + 12\% on ScoreNet) and invasive cancer (94.6\%). Our model showed strong performance in clinically critical categories such as FEA (80.3\%) and DCIS (67.1\%), which are often difficult to differentiate due to subtle architectural differences.
Compared to cell-graph-based methods (CGC-Net, Patch-GNN, CG-GNN, and HACT-Net), our approach showed consistent improvements, particularly for UDH and ADH categories that depend on subtle cytological features. Against MIL-based approaches (CLAM and TransMIL), it better captures spatial relationships between cells and tissue microenvironments. Outperforming the ScoreNet suggests that explicitly modeling tissue-cell interactions provides more discriminative features.
\begin{table}
\begin{center}
\scriptsize
\setlength{\tabcolsep}{4pt}
\begin{tabular}{l|c|c|c|c|c|c|c|c}
\hline
\textbf{Model} & \textbf{Normal} & \textbf{Benign} & \textbf{UDH} & \textbf{ADH} & \textbf{FEA} & \textbf{DCIS} & \textbf{IC} & \textbf{Total}\\
\hline
CGC-Net\cite{zhou2019cgcgnn} & 30.8 ± 5.3 & 31.6 ± 4.7 & 17.3 ± 3.4 & 24.5 ± 5.2 & 59.0 ± 3.6 & 49.4 ± 3.4 & 75.3 ± 3.2 & 43.6 ± 0.5 \\
Patch-GNN\cite{aygunecs2020graphpatchgnn} & 52.5 ± 3.3 & 47.6 ± 2.2 & 23.7 ± 4.6 & 30.7 ± 1.8 & 60.7 ± 5.3 & 58.8 ± 1.1 & 81.6 ± 2.2 & 52.1 ± 0.6 \\
CG-GNN\cite{pati2020hact} & 63.6 ± 4.9 & 47.7 ± 3.1 & 34.7 ± 4.9 & 28.5 ± 4.3 & 72.1 ± 3.6 & 54.6 ± 3.2 & 82.2 ± 4.0 & 56.6 ± 1.3 \\
HACT-Net\cite{pati2020hact} & 61.6 ± 2.1 & 47.5 ± 2.9 & 43.6 ± 1.9 & 40.4 ± 2.5 & 74.2 ± 4.6 & \underline{66.4 ± 3.6} & 88.4 ± 0.2 & 61.5 ± 0.9 \\
\hline
CLAM\cite{Lu2021_CLAM} & 59.4 ± 2.0 & 47.7 ± 1.2 & 31.7 ± 0.7 & 20.1 ± 3.4 & 68.3 ± 4.0 & 59.9 ± 1.7 & 86.8 ± 3.6 & 54.8 ± 1.0 \\
TransMIL\cite{shao2021transmil} & 47.6 ± 9.8 & 42.9 ± 3.6 & 31.5 ± 5.3 & 38.4 ± 5.9 & 72.7 ± 3.6 & 62.7 ± 2.9 & 87.1 ± 3.9 & 57.5 ± 0.7 \\
ScoreNet\cite{Thomas2023_WACV} & \underline{64.3 ± 1.5} & \textbf{54.0 ± 2.2} & \underline{45.3 ± 3.4} & \textbf{46.7 ± 1.0} & \underline{78.1 ± 2.8} & 62.9 ± 2.0 & \underline{91.0 ± 1.4} & \underline{64.4 ± 0.9} \\
\hline
\textbf{Ours} & \textbf{76.3 ± 3.7} & \underline{51.6 ± 1.5} & \textbf{47.5 ± 2.8} & \underline{45.0 ± 2.6} & \textbf{80.3 ± 3.5} & \textbf{67.1 ± 2.3} & \textbf{94.6 ± 1.2} & \textbf{66.5 ± 0.8} \\
\hline
\end{tabular}
\end{center}
\caption{Comparison with the prior art for breast cancer subtyping on the BRACS dataset, including the F1 score for each category and the weighted F1 score for seven-category classification. The results are presented in percentages(\%). The
best results are highlighted in \textbf{bold}, and the second-best results are \underline{underlined}.}
\label{tab:main}
\end{table}

\begin{table}[t]
\setlength{\tabcolsep}{1mm}
\begin{center}
\scriptsize
\begin{tabular}{l|c|c|c|c|c|c|c}
\hline
\textbf{Aggregator} & \textbf{Complexity} & \textbf{GPU Mem} & \textbf{Latency} & \textbf{Throughput} & \textbf{GFLOPs} & \textbf{Speed-up} & \textbf{Weighted F1} \\
& & (GB) & (ms) & (patch/s) & (G) & (↑) & (\%) \\
\hline
Self-Attention\cite{vaswani2017attention}          & $\mathcal{O}(n^{2})$ & 5.2 & 42.8 & 23  &10.52 & 1.0$\times$ & 64.2 \\
Set Transformer\cite{lee2019set}         & $\mathcal{O}(n^{2})$ & 4.8 & 38.4 & 26  & 9.47  & 1.1$\times$ & 65.7 \\
DeepSets\cite{zaheer2017deep} (mean)         & $\mathcal{O}(n)$     & 2.1 & 15.6 & 64  &3.82  & 2.8$\times$ & 62.3 \\
\textbf{Aggr-RWKV (ours)} & $\mathcal{O}(n)$     & 2.4 & 16.8 & 71 & 4.09  & 3.1$\times$ & 66.5 \\
\hline
\end{tabular}
\end{center}
\caption{Comparison of cell aggregation methods, including computational efficiency on BRACS validation set ($n\!=\!512$ nuclei per patch) and Total Weighted F1 score achieved by complete dual-stream model on BRACS test set when using each aggregator (ablation study). Computational metrics are averaged over 1,000 forward passes on a single NVIDIA A100.}
\label{tab:aggregator_comparison}
\end{table}

\begin{figure*}
\centering
\includegraphics[width=.9\linewidth]{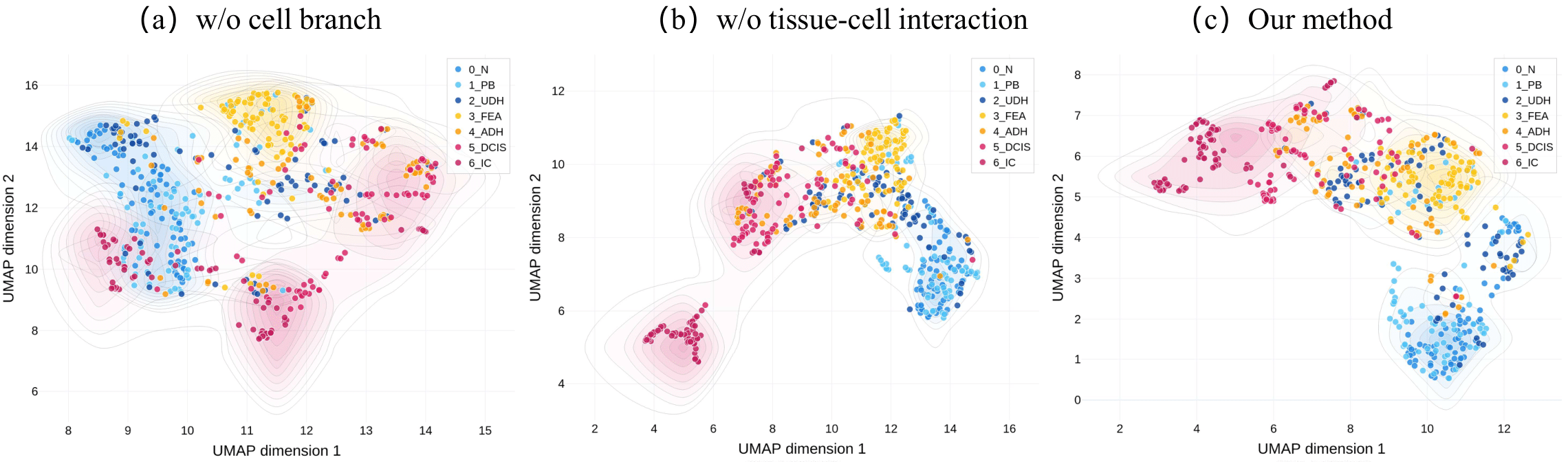}
\caption{UMAP visualization of feature embeddings from the ablation study: (a) without cell branch, (b) without tissue–cell interaction (simple concatenation), and (c) full model.}
\label{fig:fig2}
\end{figure*}
% \begin{table}
% \begin{center}
% \begin{tabular}{|l|c|}
% \hline
% Model & BRACS → BACH \\
% \hline\hline
% HACT-Net & 40.2±2.8 \\
% CLAM & 57.5±3.6 \\
% TransMIL & 46.5±10.2 \\
% HMAE & 67.3±3.2 \\
% \hline
% Ours & \textbf{70.2±4.7} \\
% \hline
% \end{tabular}
% \end{center}
% \caption{Cross-dataset generalization performance. Models were trained on BRACS and evaluated on BACH using a linear classifier on frozen feature embeddings.}
% \end{table}

\noindent\textbf{Efficiency analysis:} Table~\ref{tab:aggregator_comparison} compares different cell aggregation methods in terms of computational efficiency and diagnostic accuracy. Quadratic complexity methods (Self-Attention and Set Transformer) achieve good accuracy but with high computational costs. DeepSets offers linear complexity and better efficiency but lower accuracy (62.3\% F1). Our Aggr-RWKV achieves both linear complexity and the highest accuracy (66.5\% F1) while maintaining excellent computational efficiency (3.1$\times$ speed-up over Self-Attention), making it ideal for high-throughput clinical applications.
% \subsection{Generalization Capabilities}

\noindent\textbf{Generalization capabilities:} To evaluate generalization capability, we assessed cross-dataset performance by transferring models trained on BRACS to the BACH dataset. Our method achieved 70.2±4.7\% F1 score on BACH without fine-tuning, significantly outperforming other methods including HMAE \cite{chiocchetti2024beyond} (67.3±3.2\%), CLAM \cite{Lu2021_CLAM} (57.5±3.6\%), Trans-MIL \cite{shao2021transmil} (46.5±10.2\%), and HACT-Net \cite{pati2020hact} (40.2±2.8\%). This strong cross-dataset performance indicates that the proposed method captures fundamental histopathological patterns that transfer well across datasets, despite variations in image acquisition and annotation protocols.
To further assess domain generalization, we train the model on the UHU training set and evaluate it in-domain on UHU test set and cross-domain on UBC dataset without fine-tuning. As shown in Table~\ref{tab:prostate_results}, ITC-RWKV achieves strong in-domain results and markedly outperforms all baselines on the more challenging cross-domain test, demonstrating robust and transferable representations. These compelling results, across both breast and prostate cancer, underscore the model's ability to learn robust, transferable features, confirming its strong generalization capability for broader clinical applications.  

\begin{table}

\begin{center}

\scriptsize

\setlength{\tabcolsep}{4pt}

\begin{tabular}{l|c|c|c|c}
\hline
\multirow{2}{*}{\textbf{Experiments}} & \multicolumn{2}{c|}{\textbf{In-domain}} & \multicolumn{2}{c}{\textbf{Cross-domain}} \\
\cline{2-5}
 & Acc (\%) & F1 & Acc (\%) & F1 \\
\hline
ResNet-50~\cite{he2016deep} & \underline{78.3} & 0.656 & 70.9 & \underline{0.651} \\
%ResNeXt-50~\cite{xie2017aggregated} & 76.7 & 0.634 & 70.7 & \textbf{0.670} \\
ViT-B32~\cite{dosovitskiy2020imagevit}& 77.4 & \underline{0.665}  & \textbf{72.4} & 0.637 \\
CTransPath~\cite{wang2022transformer} & 64.5 & 0.643 & 61.1 & 0.614 \\
DCAH-Net~\cite{zhang2023hepatocellulardcah} & 54.1 & 0.420 & 61.9 & 0.526 \\
HiFuse~\cite{huo2024hifuse} & 62.7  & 0.457 & 61.2 & 0.501 \\
\hline
\textbf{ITC-RWKV (ours)} & \textbf{79.5} & \textbf{0.673} & \underline{72.1} & \textbf{0.662} \\
\hline
\end{tabular}
\end{center}
\caption{Generalization performance on two prostate cancer test sets. We report Accuracy (Acc) and F1-score (F1). Best results are highlighted in bold.}
\label{tab:prostate_results}
\end{table}

% \subsection{Feature Representation Analysis}
\noindent\textbf{Ablation study:} To visualize feature space organization, we project embeddings to 2D with UMAP (Fig.~\ref{fig:fig2}). Without the cell branch (Fig.~\ref{fig:fig2}a), categories dependent on nuclear morphology (ADH, DCIS) collapse into overlapping regions, as the model fails to distinguish entities with similar architecture but different cytology. With simple concatenation instead of interaction (Fig.~\ref{fig:fig2}b), separation improves but boundaries remain blurred, particularly for non-invasive lesions where spatial context determines grade. Our complete model (Fig.~\ref{fig:fig2}c) creates clear delineation: normal/benign samples form a tight cluster (blue), non-atypical proliferative lesions occupy an intermediate region (yellow), and pre-malignant/malignant classes show distinct separation (red). This progression mirrors biological continuum, confirming that our dual-stream architecture with interaction captures both architectural and cytological features essential for accurate classification.

\begin{figure*}
\centering
\includegraphics[width=.8\linewidth]{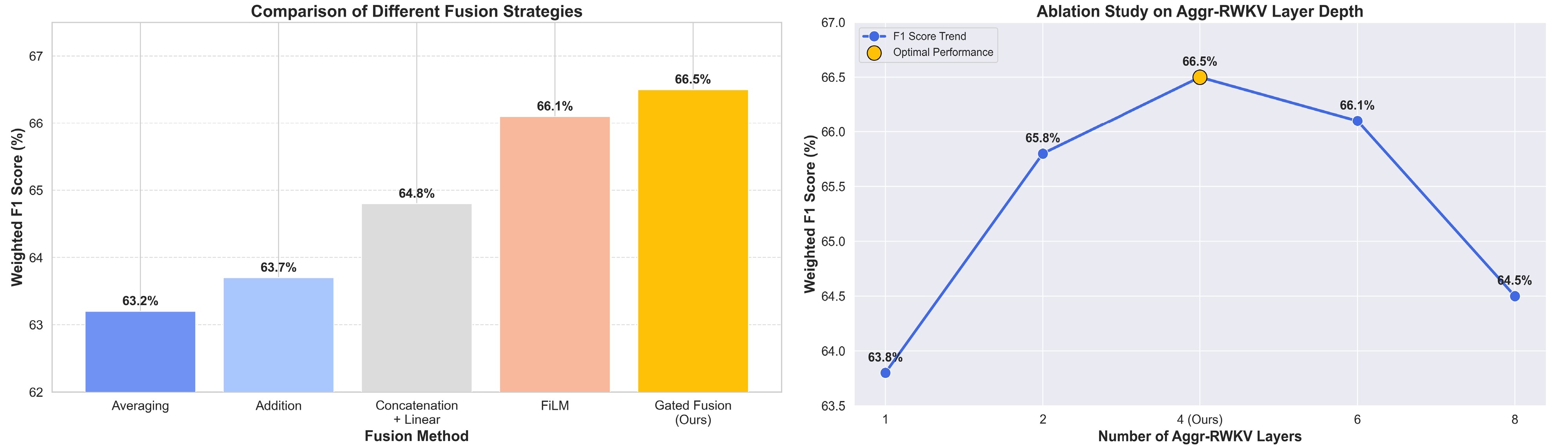}
\caption{Ablation studies on key model components on the BRACS dataset. Left: Comparison of different fusion strategies for combining tissue and cell-level features. Right: Impact of varying the layer depth of the Aggr-RWKV module.}
\label{fig:abl}
\end{figure*}

In addition to this qualitative analysis, we conducted quantitative ablation studies to rigorously validate our key architectural choices (in Fig.~\ref{fig:abl}). We compared our gated fusion against simpler strategies (e.g., averaging, addition) and FiLM, a method that uses one feature stream to apply a learned affine transformation to the other. Our proposed gated fusion achieved the highest F1 score, demonstrating its superior capability for adaptive feature integration. Concurrently, we analyzed the depth of the Aggr-RWKV module and observed that model performance peaked at 4 layers, indicating an optimal balance between representation power and generalization. Together, these qualitative and quantitative results affirm the robustness and efficacy of our model's core designs.

\subsection{Interpretability}
% \noindent\textbf{Interpretability}
To provide interpretability for our model's decisions, we generate tissue region importance maps highlighting areas most critical for classification. The computation process involves: (1) calculating cell influence maps based on attention weights from our model, (2) applying Gaussian smoothing ($\sigma=15$) to create continuous importance regions, (3) normalizing values to 0-1 range, (4) converting to a color map using the viridis color scheme, and (5) overlaying on the original image with 0.6 alpha transparency. This process creates intuitive visualizations of regions that most influenced the model's diagnostic decisions.

\begin{figure*}
\centering
\includegraphics[width=.8\linewidth]{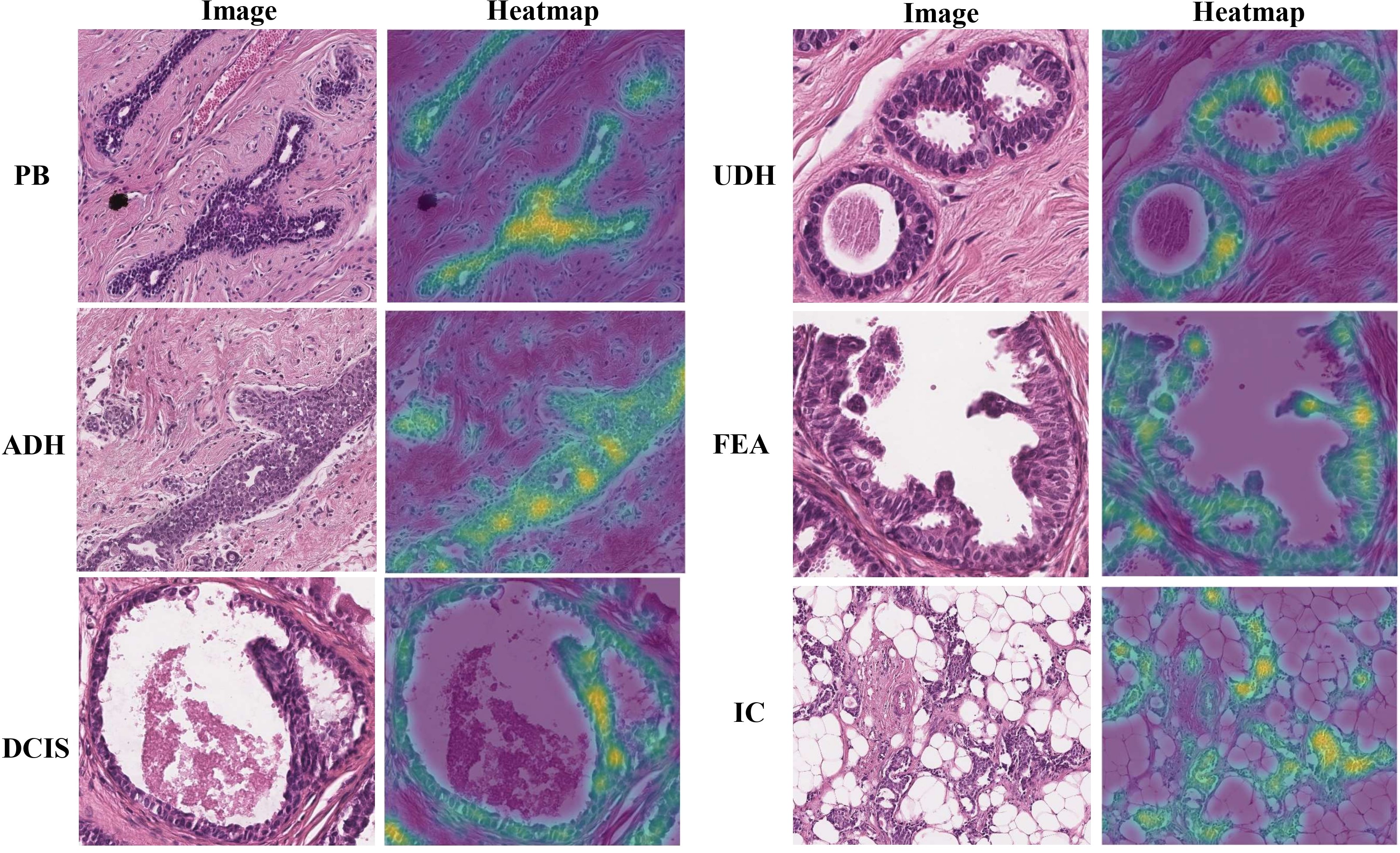}
\caption{Tissue region importance heatmaps across different breast pathologies. For each pair, the left shows original H\&E images and the right shows corresponding importance maps where yellow-green highlights indicate regions most influential for classification.}
\label{fig:fig3}
\end{figure*}

The resulting heatmaps reveal diagnostically relevant patterns across different breast pathologies (Fig.~\ref{fig:fig3}). In benign proliferative lesions (PB), strong signals are concentrated in the hyperplastic areas of the ductal epithelium and its interface with the stroma; In Usual Ductal Hyperplasia (UDH), attention is focused on typical areas with irregular cell arrangement, thickened epithelial layers, and papillary projections extending into the lumen; ADH heatmaps accurately cover areas exhibiting cellular atypia, with the highest intensity corresponding to the most prominent cytological abnormalities; Flat Epithelial Atypia (FEA) appears as a ring-shaped high intensity surrounding dilated ducts, precisely indicating the flattened epithelial cells lining the ducts. For malignant lesions, the model's interpretability is particularly prominent: In Ductal Carcinoma In Situ (DCIS), heatmaps prominently mark the most significant cellular atypia and the interface between comedo-type necrosis and viable cells; Whereas in invasive carcinoma (IC), the heatmaps simultaneously present four clinically relevant features: the invasive front (tumor-fat interface), differences between the center and periphery of tumor nests, strong signals at the tumor-stroma interface, and scattered small tumor clusters within adipose tissue, which highly align with known tumor heterogeneity and invasion patterns, suggesting that the model captures crucial prognostic information beyond the classification task.

\section{Conclusion}
The proposed dual-stream framework, together with an Aggr-RWKV module and the tissue-cell interaction mechanism, offers an effective strategy for bridging micro- and macro-level reasoning in histopathological image analysis. The scalability of Aggr-RWKV and the flexibility of the interaction module make the framework well-suited for dense cellular environments and diverse diagnostic scenarios. These findings suggest that incorporating structured, multilevel representations can meaningfully advance fine-grained classification performance and pave the way for more interpretable and generalizable computational pathology systems. Future work will focus on extending the framework to whole-slide inference and integrating spatial priors or multimodal clinical context.
%-------------------------------------------------------------------------

\bibliography{bmvc_final}

\begin{thebibliography}{44}
\providecommand{\natexlab}[1]{#1}
\providecommand{\url}[1]{\texttt{#1}}
\expandafter\ifx\csname urlstyle\endcsname\relax
  \providecommand{\doi}[1]{doi: #1}\else
  \providecommand{\doi}{doi: \begingroup \urlstyle{rm}\Url}\fi

\bibitem[Allemani et~al.(2015)Allemani, Weir, Carreira, Harewood, Spika, Wang, Bannon, Ahn, Johnson, Bonaventure, Marcos-Gragera, Stiller, Azevedo~e Silva, Chen, Ogunbiyi, Rachet, Soeberg, You, Matsuda, Bielska-Lasota, Storm, Tucker, and Coleman]{Allemani2015_CONCORD2}
Claudia Allemani, Hannah~K. Weir, Helena Carreira, Rebecca Harewood, Dominik Spika, Xue~S. Wang, Fiona Bannon, James~V. Ahn, Christopher~J. Johnson, Audrey Bonaventure, Rafael Marcos-Gragera, Charles Stiller, Gulnar Azevedo~e Silva, Wan-Qing Chen, Olufunmilayo~J. Ogunbiyi, Bernard Rachet, Matthew~J. Soeberg, He~You, Tomohiro Matsuda, Magdalena Bielska-Lasota, Hans Storm, Thomas~C. Tucker, and Michel~P. Coleman.
\newblock Global surveillance of cancer survival 1995--2009: analysis of individual data for 25,676,887 patients from 279 population-based registries in 67 countries ({CONCORD{-}2}).
\newblock \emph{The Lancet}, 385\penalty0 (9972):\penalty0 977--1010, March 2015.
\newblock \doi{10.1016/S0140-6736(14)62038-9}.
\newblock Epub 2014 Nov 26; erratum in \textit{Lancet}. 2015 Mar 14;385(9972):946.

\bibitem[Aresta et~al.(2019)Aresta, Ara{\'u}jo, Kwok, Chennamsetty, Safwan, Alex, Marami, Prastawa, Chan, Donovan, et~al.]{aresta2019bach}
Guilherme Aresta, Teresa Ara{\'u}jo, Scotty Kwok, Sai~Saketh Chennamsetty, Mohammed Safwan, Varghese Alex, Bahram Marami, Marcel Prastawa, Monica Chan, Michael Donovan, et~al.
\newblock Bach: Grand challenge on breast cancer histology images.
\newblock \emph{Medical image analysis}, 56:\penalty0 122--139, 2019.

\bibitem[Arvaniti et~al.(2018)Arvaniti, Fricker, Moret, Rupp, Hermanns, Fankhauser, Wey, Wild, Rueschoff, and Claassen]{arvaniti2018automated}
Eirini Arvaniti, Kim~S Fricker, Michael Moret, Niels Rupp, Thomas Hermanns, Christian Fankhauser, Norbert Wey, Peter~J Wild, Jan~H Rueschoff, and Manfred Claassen.
\newblock Automated gleason grading of prostate cancer tissue microarrays via deep learning.
\newblock \emph{Scientific reports}, 8\penalty0 (1):\penalty0 12054, 2018.

\bibitem[Ayg{\"u}ne{\c{s}} et~al.(2020)Ayg{\"u}ne{\c{s}}, Aksoy, Cinbi{\c{s}}, K{\"o}semehmeto{\u{g}}lu, {\"O}nder, and {\"U}ner]{aygunecs2020graphpatchgnn}
Bulut Ayg{\"u}ne{\c{s}}, Selim Aksoy, Ramazan~G{\"o}kberk Cinbi{\c{s}}, Kemal K{\"o}semehmeto{\u{g}}lu, Sevgen {\"O}nder, and Ay{\c{s}}eg{\"u}l {\"U}ner.
\newblock Graph convolutional networks for region of interest classification in breast histopathology.
\newblock In \emph{Medical Imaging 2020: Digital Pathology}, volume 11320, pages 134--141. SPIE, 2020.

\bibitem[Brancati et~al.(2022)Brancati, Anniciello, Pati, Riccio, Scognamiglio, Jaume, De~Pietro, Di~Bonito, Foncubierta, Botti, et~al.]{brancati2022bracs}
Nadia Brancati, Anna~Maria Anniciello, Pushpak Pati, Daniel Riccio, Giosu{\`e} Scognamiglio, Guillaume Jaume, Giuseppe De~Pietro, Maurizio Di~Bonito, Antonio Foncubierta, Gerardo Botti, et~al.
\newblock Bracs: A dataset for breast carcinoma subtyping in h\&e histology images.
\newblock \emph{Database}, 2022:\penalty0 baac093, 2022.

\bibitem[Campanella et~al.(2019)Campanella, Hanna, Geneslaw, Miraflor, Werneck Krauss~Silva, Busam, Brogi, Reuter, Klimstra, and Fuchs]{Campanella2019_CompPathology}
Gabriele Campanella, Matthew~G. Hanna, Luke Geneslaw, Allen Miraflor, Vitor Werneck Krauss~Silva, Klaus~J. Busam, Edi Brogi, Victor~E. Reuter, David~S. Klimstra, and Thomas~J. Fuchs.
\newblock Clinical-grade computational pathology using weakly supervised deep learning on whole slide images.
\newblock \emph{Nature Medicine}, 25\penalty0 (8):\penalty0 1301--1309, 2019.
\newblock ISSN 1546-170X.
\newblock \doi{10.1038/s41591-019-0508-1}.
\newblock URL \url{https://doi.org/10.1038/s41591-019-0508-1}.

\bibitem[Chen et~al.(2021)Chen, Lu, Shaban, Chen, Chen, Williamson, and Mahmood]{chen2021patchgcn}
Richard~J Chen, Ming~Y Lu, Muhammad Shaban, Chengkuan Chen, Tiffany~Y Chen, Drew~FK Williamson, and Faisal Mahmood.
\newblock Whole slide images are 2d point clouds: Context-aware survival prediction using patch-based graph convolutional networks.
\newblock In \emph{Medical Image Computing and Computer Assisted Intervention--MICCAI 2021: 24th International Conference, Strasbourg, France, September 27--October 1, 2021, Proceedings, Part VIII 24}, pages 339--349. Springer, 2021.

\bibitem[Chen et~al.(2024)Chen, Ding, Lu, Williamson, Jaume, Song, Chen, Zhang, Shao, Shaban, Williams, Oldenburg, Weishaupt, Wang, Vaidya, Le, Gerber, Sahai, Williams, and Mahmood]{Chen2024_UNI}
Richard~J. Chen, Tong Ding, Ming~Y. Lu, Drew F.~K. Williamson, Guillaume Jaume, Andrew~H. Song, Bowen Chen, Andrew Zhang, Daniel Shao, Muhammad Shaban, Mane Williams, Lukas Oldenburg, Luca~L. Weishaupt, Judy~J. Wang, Anurag Vaidya, Long~Phi Le, Georg Gerber, Sharifa Sahai, Walt Williams, and Faisal Mahmood.
\newblock Towards a general-purpose foundation model for computational pathology.
\newblock \emph{Nature Medicine}, 30\penalty0 (3):\penalty0 850--862, 2024.
\newblock ISSN 1546-170X.
\newblock \doi{10.1038/s41591-024-02857-3}.
\newblock URL \url{https://doi.org/10.1038/s41591-024-02857-3}.

\bibitem[Chiocchetti et~al.(2024)Chiocchetti, Dossena, Irwin, and Portinale]{chiocchetti2024beyond}
Annalisa Chiocchetti, Marco Dossena, Christopher Irwin, and Luigi Portinale.
\newblock Beyond labels: A self-supervised framework with masked autoencoders and random cropping for breast cancer subtype classification.
\newblock \emph{arXiv preprint arXiv:2410.12006}, 2024.

\bibitem[Ciga et~al.(2022)Ciga, Xu, and Martel]{ciga2022self}
Ozan Ciga, Tony Xu, and Anne~Louise Martel.
\newblock Self supervised contrastive learning for digital histopathology.
\newblock \emph{Machine learning with applications}, 7:\penalty0 100198, 2022.

\bibitem[Dosovitskiy et~al.(2020)Dosovitskiy, Beyer, Kolesnikov, Weissenborn, Zhai, Unterthiner, Dehghani, Minderer, Heigold, Gelly, et~al.]{dosovitskiy2020imagevit}
Alexey Dosovitskiy, Lucas Beyer, Alexander Kolesnikov, Dirk Weissenborn, Xiaohua Zhai, Thomas Unterthiner, Mostafa Dehghani, Matthias Minderer, Georg Heigold, Sylvain Gelly, et~al.
\newblock An image is worth 16x16 words: Transformers for image recognition at scale.
\newblock \emph{arXiv preprint arXiv:2010.11929}, 2020.

\bibitem[Duan et~al.(2024)Duan, Wang, Chen, Zhu, Lu, Lu, Qiao, Li, Dai, and Wang]{duan2024visionrwkv}
Yuchen Duan, Weiyun Wang, Zhe Chen, Xizhou Zhu, Lewei Lu, Tong Lu, Yu~Qiao, Hongsheng Li, Jifeng Dai, and Wenhai Wang.
\newblock Vision-rwkv: Efficient and scalable visual perception with rwkv-like architectures.
\newblock \emph{arXiv preprint arXiv:2403.02308}, 2024.

\bibitem[Fei et~al.(2024)Fei, Fan, Yu, Li, and Huang]{fei2024diffusionrwkv}
Zhengcong Fei, Mingyuan Fan, Changqian Yu, Debang Li, and Junshi Huang.
\newblock Diffusion-rwkv: Scaling rwkv-like architectures for diffusion models.
\newblock \emph{arXiv preprint arXiv:2404.04478}, 2024.

\bibitem[Gamper et~al.(2020)Gamper, Koohbanani, Graham, Jahanifar, Khurram, Azam, Hewitt, and Rajpoot]{gamper2020pannuke}
Jevgenij Gamper, Navid~Alemi Koohbanani, Simon Graham, Mostafa Jahanifar, Syed~Ali Khurram, Ayesha Azam, Katherine Hewitt, and Nasir Rajpoot.
\newblock Pannuke dataset extension, insights and baselines.
\newblock \emph{arXiv preprint arXiv:2003.10778}, 2020.

\bibitem[Graham et~al.(2019)Graham, Vu, Raza, Azam, Tsang, Kwak, and Rajpoot]{graham2019hover}
Simon Graham, Quoc~Dang Vu, Shan E~Ahmed Raza, Ayesha Azam, Yee~Wah Tsang, Jin~Tae Kwak, and Nasir Rajpoot.
\newblock Hover-net: Simultaneous segmentation and classification of nuclei in multi-tissue histology images.
\newblock \emph{Medical image analysis}, 58:\penalty0 101563, 2019.

\bibitem[Gu et~al.(2024)Gu, Yang, An, Feng, Liu, Cai, and Deng]{gu2024rwkvclip}
Tiancheng Gu, Kaicheng Yang, Xiang An, Ziyong Feng, Dongnan Liu, Weidong Cai, and Jiankang Deng.
\newblock Rwkv-clip: a robust vision-language representation learner.
\newblock \emph{arXiv preprint arXiv:2406.06973}, 2024.

\bibitem[He et~al.(2016)He, Zhang, Ren, and Sun]{he2016deep}
Kaiming He, Xiangyu Zhang, Shaoqing Ren, and Jian Sun.
\newblock Deep residual learning for image recognition.
\newblock In \emph{Proceedings of the IEEE conference on computer vision and pattern recognition}, pages 770--778, 2016.

\bibitem[He et~al.(2025)He, Zhang, Peng, He, Li, Wang, and Wang]{he2025pointrwkv}
Qingdong He, Jiangning Zhang, Jinlong Peng, Haoyang He, Xiangtai Li, Yabiao Wang, and Chengjie Wang.
\newblock Pointrwkv: Efficient rwkv-like model for hierarchical point cloud learning.
\newblock In \emph{Proceedings of the AAAI Conference on Artificial Intelligence}, volume~39, pages 3410--3418, 2025.

\bibitem[H{\"o}rst et~al.(2024)H{\"o}rst, Rempe, Heine, Seibold, Keyl, Baldini, Ugurel, Siveke, Gr{\"u}nwald, Egger, et~al.]{horst2024cellvit}
Fabian H{\"o}rst, Moritz Rempe, Lukas Heine, Constantin Seibold, Julius Keyl, Giulia Baldini, Selma Ugurel, Jens Siveke, Barbara Gr{\"u}nwald, Jan Egger, et~al.
\newblock Cellvit: Vision transformers for precise cell segmentation and classification.
\newblock \emph{Medical Image Analysis}, 94:\penalty0 103143, 2024.

\bibitem[Huo et~al.(2024)Huo, Sun, Tian, Wang, Yu, Long, Zhang, and Li]{huo2024hifuse}
Xiangzuo Huo, Gang Sun, Shengwei Tian, Yan Wang, Long Yu, Jun Long, Wendong Zhang, and Aolun Li.
\newblock Hifuse: Hierarchical multi-scale feature fusion network for medical image classification.
\newblock \emph{Biomedical Signal Processing and Control}, 87:\penalty0 105534, 2024.

\bibitem[Ilse et~al.(2018)Ilse, Tomczak, and Welling]{ilse2018attentionmil}
Maximilian Ilse, Jakub Tomczak, and Max Welling.
\newblock Attention-based deep multiple instance learning.
\newblock In \emph{International conference on machine learning}, pages 2127--2136. PMLR, 2018.

\bibitem[Karimi et~al.(2019)Karimi, Nir, Fazli, Black, Goldenberg, and Salcudean]{karimi2019deep}
Davood Karimi, Guy Nir, Ladan Fazli, Peter~C Black, Larry Goldenberg, and Septimiu~E Salcudean.
\newblock Deep learning-based gleason grading of prostate cancer from histopathology images—role of multiscale decision aggregation and data augmentation.
\newblock \emph{IEEE journal of biomedical and health informatics}, 24\penalty0 (5):\penalty0 1413--1426, 2019.

\bibitem[Kather et~al.(2019)Kather, Pearson, Halama, Jäger, Krause, Loosen, Marx, Boor, Tacke, Neumann, Grabsch, Yoshikawa, Brenner, Chang-Claude, Hoffmeister, Trautwein, and Luedde]{Kather2019_MSI}
Jakob~Nikolas Kather, Alexander~T. Pearson, Niels Halama, Dirk Jäger, Jeremias Krause, Sven~H. Loosen, Alexander Marx, Peter Boor, Frank Tacke, Ulf~Peter Neumann, Heike~I. Grabsch, Takaki Yoshikawa, Hermann Brenner, Jenny Chang-Claude, Michael Hoffmeister, Christian Trautwein, and Tom Luedde.
\newblock Deep learning can predict microsatellite instability directly from histology in gastrointestinal cancer.
\newblock \emph{Nature Medicine}, 25\penalty0 (7):\penalty0 1054--1056, 2019.
\newblock ISSN 1546-170X.
\newblock \doi{10.1038/s41591-019-0462-y}.
\newblock URL \url{https://doi.org/10.1038/s41591-019-0462-y}.

\bibitem[Kingma(2014)]{kingma2014adam}
Diederik~P Kingma.
\newblock Adam: A method for stochastic optimization.
\newblock \emph{arXiv preprint arXiv:1412.6980}, 2014.

\bibitem[Koohbanani et~al.(2021)Koohbanani, Unnikrishnan, Khurram, Krishnaswamy, and Rajpoot]{koohbanani2021self}
Navid~Alemi Koohbanani, Balagopal Unnikrishnan, Syed~Ali Khurram, Pavitra Krishnaswamy, and Nasir Rajpoot.
\newblock Self-path: Self-supervision for classification of pathology images with limited annotations.
\newblock \emph{IEEE Transactions on Medical Imaging}, 40\penalty0 (10):\penalty0 2845--2856, 2021.

\bibitem[Lee et~al.(2019)Lee, Lee, Kim, Kosiorek, Choi, and Teh]{lee2019set}
Juho Lee, Yoonho Lee, Jungtaek Kim, Adam Kosiorek, Seungjin Choi, and Yee~Whye Teh.
\newblock Set transformer: A framework for attention-based permutation-invariant neural networks.
\newblock In \emph{International conference on machine learning}, pages 3744--3753. PMLR, 2019.

\bibitem[Li et~al.(2021)Li, Li, and Eliceiri]{li2021dualself}
Bin Li, Yin Li, and Kevin~W Eliceiri.
\newblock Dual-stream multiple instance learning network for whole slide image classification with self-supervised contrastive learning.
\newblock In \emph{Proceedings of the IEEE/CVF conference on computer vision and pattern recognition}, pages 14318--14328, 2021.

\bibitem[Lu et~al.(2021)Lu, Williamson, Chen, Chen, Barbieri, and Mahmood]{Lu2021_CLAM}
Ming~Y. Lu, Drew F.~K. Williamson, Tiffany~Y. Chen, Richard~J. Chen, Matteo Barbieri, and Faisal Mahmood.
\newblock Data-efficient and weakly supervised computational pathology on whole-slide images.
\newblock \emph{Nature Biomedical Engineering}, 5\penalty0 (6):\penalty0 555--570, 2021.
\newblock ISSN 2157-846X.
\newblock \doi{10.1038/s41551-020-00682-w}.
\newblock URL \url{https://doi.org/10.1038/s41551-020-00682-w}.

\bibitem[Lu et~al.(2024)Lu, Chen, Williamson, Chen, Liang, Ding, Jaume, Odintsov, Le, Gerber, Parwani, Zhang, and Mahmood]{Lu2024_CONCH}
Ming~Y. Lu, Bowen Chen, Drew F.~K. Williamson, Richard~J. Chen, Ivy Liang, Tong Ding, Guillaume Jaume, Igor Odintsov, Long~Phi Le, Georg Gerber, Anil~V. Parwani, Andrew Zhang, and Faisal Mahmood.
\newblock A visual-language foundation model for computational pathology.
\newblock \emph{Nature Medicine}, 30\penalty0 (3):\penalty0 863--874, 2024.
\newblock ISSN 1546-170X.
\newblock \doi{10.1038/s41591-024-02856-4}.
\newblock URL \url{https://doi.org/10.1038/s41591-024-02856-4}.

\bibitem[Ma et~al.(2024)Ma, Guo, Zhou, Wang, Xu, Cai, Zhu, Jin, Lin, Jiang, et~al.]{ma2024gpfm}
Jiabo Ma, Zhengrui Guo, Fengtao Zhou, Yihui Wang, Yingxue Xu, Yu~Cai, Zhengjie Zhu, Cheng Jin, Yi~Lin, Xinrui Jiang, et~al.
\newblock Towards a generalizable pathology foundation model via unified knowledge distillation.
\newblock \emph{arXiv preprint arXiv:2407.18449}, 2024.

\bibitem[Pati et~al.(2020)Pati, Jaume, Fernandes, Foncubierta-Rodr{\'\i}guez, Feroce, Anniciello, Scognamiglio, Brancati, Riccio, Di~Bonito, et~al.]{pati2020hact}
Pushpak Pati, Guillaume Jaume, Lauren~Alisha Fernandes, Antonio Foncubierta-Rodr{\'\i}guez, Florinda Feroce, Anna~Maria Anniciello, Giosue Scognamiglio, Nadia Brancati, Daniel Riccio, Maurizio Di~Bonito, et~al.
\newblock Hact-net: A hierarchical cell-to-tissue graph neural network for histopathological image classification.
\newblock In \emph{Uncertainty for Safe Utilization of Machine Learning in Medical Imaging, and Graphs in Biomedical Image Analysis: Second International Workshop, UNSURE 2020, and Third International Workshop, GRAIL 2020, Held in Conjunction with MICCAI 2020, Lima, Peru, October 8, 2020, Proceedings 2}, pages 208--219. Springer, 2020.

\bibitem[Pati et~al.(2022)Pati, Jaume, Foncubierta-Rodriguez, Feroce, Anniciello, Scognamiglio, Brancati, Fiche, Dubruc, Riccio, et~al.]{pati2022hierarchicalgnn}
Pushpak Pati, Guillaume Jaume, Antonio Foncubierta-Rodriguez, Florinda Feroce, Anna~Maria Anniciello, Giosue Scognamiglio, Nadia Brancati, Maryse Fiche, Estelle Dubruc, Daniel Riccio, et~al.
\newblock Hierarchical graph representations in digital pathology.
\newblock \emph{Medical image analysis}, 75:\penalty0 102264, 2022.

\bibitem[Peng et~al.(2023)Peng, Alcaide, Anthony, Albalak, Arcadinho, Biderman, Cao, Cheng, Chung, Grella, et~al.]{peng2023rwkv}
Bo~Peng, Eric Alcaide, Quentin Anthony, Alon Albalak, Samuel Arcadinho, Stella Biderman, Huanqi Cao, Xin Cheng, Michael Chung, Matteo Grella, et~al.
\newblock Rwkv: Reinventing rnns for the transformer era.
\newblock \emph{arXiv preprint arXiv:2305.13048}, 2023.

\bibitem[Shao et~al.(2021)Shao, Bian, Chen, Wang, Zhang, Ji, et~al.]{shao2021transmil}
Zhuchen Shao, Hao Bian, Yang Chen, Yifeng Wang, Jian Zhang, Xiangyang Ji, et~al.
\newblock Transmil: Transformer based correlated multiple instance learning for whole slide image classification.
\newblock \emph{Advances in neural information processing systems}, 34:\penalty0 2136--2147, 2021.

\bibitem[Stegmüller et~al.(2023)Stegmüller, Bozorgtabar, Spahr, and Thiran]{Thomas2023_WACV}
Thomas Stegmüller, Behzad Bozorgtabar, Antoine Spahr, and Jean-Philippe Thiran.
\newblock Scorenet: Learning non-uniform attention and augmentation for transformer-based histopathological image classification.
\newblock In \emph{2023 IEEE/CVF Winter Conference on Applications of Computer Vision (WACV)}, pages 6159--6168, 2023.
\newblock \doi{10.1109/WACV56688.2023.00611}.

\bibitem[Tosta et~al.(2019)Tosta, de~Faria, Neves, and do~Nascimento]{tosta2019computational}
Tha{\'\i}na A~Azevedo Tosta, Paulo~Rog{\'e}rio de~Faria, Leandro~Alves Neves, and Marcelo~Zanchetta do~Nascimento.
\newblock Computational normalization of h\&e-stained histological images: Progress, challenges and future potential.
\newblock \emph{Artificial intelligence in medicine}, 95:\penalty0 118--132, 2019.

\bibitem[Vaswani et~al.(2017)Vaswani, Shazeer, Parmar, Uszkoreit, Jones, Gomez, Kaiser, and Polosukhin]{vaswani2017attention}
Ashish Vaswani, Noam Shazeer, Niki Parmar, Jakob Uszkoreit, Llion Jones, Aidan~N Gomez, {\L}ukasz Kaiser, and Illia Polosukhin.
\newblock Attention is all you need.
\newblock \emph{Advances in neural information processing systems}, 30, 2017.

\bibitem[Vorontsov et~al.(2024)Vorontsov, Bozkurt, Casson, Shaikovski, Zelechowski, Severson, Zimmermann, Hall, Tenenholtz, Fusi, et~al.]{vorontsov2024virchow}
Eugene Vorontsov, Alican Bozkurt, Adam Casson, George Shaikovski, Michal Zelechowski, Kristen Severson, Eric Zimmermann, James Hall, Neil Tenenholtz, Nicolo Fusi, et~al.
\newblock A foundation model for clinical-grade computational pathology and rare cancers detection.
\newblock \emph{Nature medicine}, 30\penalty0 (10):\penalty0 2924--2935, 2024.

\bibitem[Wang et~al.(2022)Wang, Yang, Zhang, Wang, Zhang, Yang, Huang, and Han]{wang2022transformer}
Xiyue Wang, Sen Yang, Jun Zhang, Minghui Wang, Jing Zhang, Wei Yang, Junzhou Huang, and Xiao Han.
\newblock Transformer-based unsupervised contrastive learning for histopathological image classification.
\newblock \emph{Medical image analysis}, 81:\penalty0 102559, 2022.

\bibitem[Yao et~al.(2020)Yao, Zhu, Jonnagaddala, Hawkins, and Huang]{yao2020wholeattentionmil}
Jiawen Yao, Xinliang Zhu, Jitendra Jonnagaddala, Nicholas Hawkins, and Junzhou Huang.
\newblock Whole slide images based cancer survival prediction using attention guided deep multiple instance learning networks.
\newblock \emph{Medical image analysis}, 65:\penalty0 101789, 2020.

\bibitem[Yuan et~al.(2024)Yuan, Li, Qi, Zhang, Yang, Yan, and Loy]{yuan2024samrwkv}
Haobo Yuan, Xiangtai Li, Lu~Qi, Tao Zhang, Ming-Hsuan Yang, Shuicheng Yan, and Chen~Change Loy.
\newblock Mamba or rwkv: Exploring high-quality and high-efficiency segment anything model.
\newblock \emph{arXiv preprint arXiv:2406.19369}, 2024.

\bibitem[Zaheer et~al.(2017)Zaheer, Kottur, Ravanbakhsh, Poczos, Salakhutdinov, and Smola]{zaheer2017deep}
Manzil Zaheer, Satwik Kottur, Siamak Ravanbakhsh, Barnabas Poczos, Russ~R Salakhutdinov, and Alexander~J Smola.
\newblock Deep sets.
\newblock \emph{Advances in neural information processing systems}, 30, 2017.

\bibitem[Zhang et~al.(2023)Zhang, Qiu, Li, Zhou, Hu, Weng, Sheng, Dong, and Ren]{zhang2023hepatocellulardcah}
Jinhua Zhang, Song Qiu, Qingli Li, Chenhao Zhou, Zhiqiu Hu, Jialei Weng, Xia Sheng, Qiongzhu Dong, and Ning Ren.
\newblock Hepatocellular carcinoma histopathological images grading with a novel attention-sharing hybrid network based on multi-feature fusion.
\newblock \emph{Biomedical Signal Processing and Control}, 86:\penalty0 105126, 2023.

\bibitem[Zhou et~al.(2019)Zhou, Graham, Alemi~Koohbanani, Shaban, Heng, and Rajpoot]{zhou2019cgcgnn}
Yanning Zhou, Simon Graham, Navid Alemi~Koohbanani, Muhammad Shaban, Pheng-Ann Heng, and Nasir Rajpoot.
\newblock Cgc-net: Cell graph convolutional network for grading of colorectal cancer histology images.
\newblock In \emph{Proceedings of the IEEE/CVF international conference on computer vision workshops}, pages 0--0, 2019.

\end{thebibliography}
\end{document}